\documentclass[lettersize,journal]{IEEEtran}
\usepackage{amsmath,amsfonts}
\usepackage{algorithmic}
\usepackage{algorithm}
\usepackage{cite}
\usepackage{array}
\usepackage[caption=false,font=normalsize,labelfont=sf,textfont=sf]{subfig}
\usepackage{textcomp}
\usepackage{stfloats}
\usepackage{url}
\usepackage{verbatim}
\usepackage{graphicx}
\usepackage{multirow}
\usepackage{hyperref}
\usepackage[capitalize]{cleveref}
\usepackage{cite}
\usepackage{bm} 
\usepackage[numbers]{natbib}
\usepackage{booktabs}
\usepackage{bbding}
\usepackage{tabularx} 
\usepackage{xcolor}
\usepackage{bbm}
\usepackage{hyperref}

\begin{document}

\title{RPCASSM: Robust PCA State Space Model For Infrared Small Target Detection}

\author{Pingping Liu, Aohua Li, Yubing Lu, Jin Kuang, Tongshun Zhang, Qiuzhan Zhou
\thanks{This work is supported by Jilin Province Industrial Key Core Technology Tackling Project (20230201085GX).}
\thanks{Pingping Liu , Yubing Lu and Tongshun Zhang are with the College of Computer Science and Technology, Jilin University, Changchun, Jilin, 130012, China and also with the Key Laboratory of Symbolic Computation and Knowledge Engineering of Ministry of Education, Jilin University, Changchun, Jilin 130012, China (e-mail: liupp@jlu.edu.cn; luyb24@mails.jlu.edu.cn; tszhang23@mails.jlu.edu.cn; yangming24@mails.jlu.edu.cn)}

\thanks{Aohua Li is with the College of Software, Jilin University, Changchun, 130000, Jilin, China(e-mail:liah24@mails.jlu.edu.cn).}

\thanks{Jin Kuang is with the School of Geosciences, Yangtze University, Wuhan 430100 (e-mail: gasquue@gmail.com).}

\thanks{Qiuzhan Zhou is with the College of Communication Engineering, Jilin University, Changchun, Jilin, 130012, China (e-mail: zhouqz@jlu.edu.cn)}
}

\markboth{Journal of \LaTeX\ Class Files,~Vol.~14, No.~8, August~2021}%
{Shell \MakeLowercase{\textit{et al.}}: A Sample Article Using IEEEtran.cls for IEEE Journals}

\maketitle

\begin{abstract}The detection and segmentation of infrared small targets have important application significance in the fields of surveillance and security, maritime rescue and so on. 
Due to the low occupancy of these targets in long-distance imaging, the mainstream visual state space model is inefficient and difficult to accurately model the target edge. 
The existing infrared state space models do not deviate from the mainstream visual state space structure framework from the structural properties of infrared small targets. 
In order to solve this problem, this paper proposes the RPCASSM network based on the model paradigm of robust principal component analysis(RPCA), which aims to design the background state space module(BSSM) and the target state space module(TSSM) by the nature of the infrared small target in the spatial domain. 
The BSSM aims to use the saliency of spatial heterogeneous signals to design a spatial probe scanning mechanism(SPCM) to model background information. 
The TSSM designs a deformable prompt scanning mechanism(DPCM) by using the sparsity and local highlight of the target to focus on the deformable space of the target for state space modeling. 
According to the above design, we effectively solve the problem that the existing mainstream vision state space model is difficult to accurately model the edge structure of infrared small target. 
Experimental results on the existing benchmark data sets prove the effectiveness of the RPCASSM design.
Our code will be made public at \href{https://github.com/PepperCS/RPCASSM}{RPCASSM}.
\end{abstract}

\section{Introduction}
\label{sec:intro}

\begin{figure}[t]
    \centering 
    \includegraphics[scale=0.042]{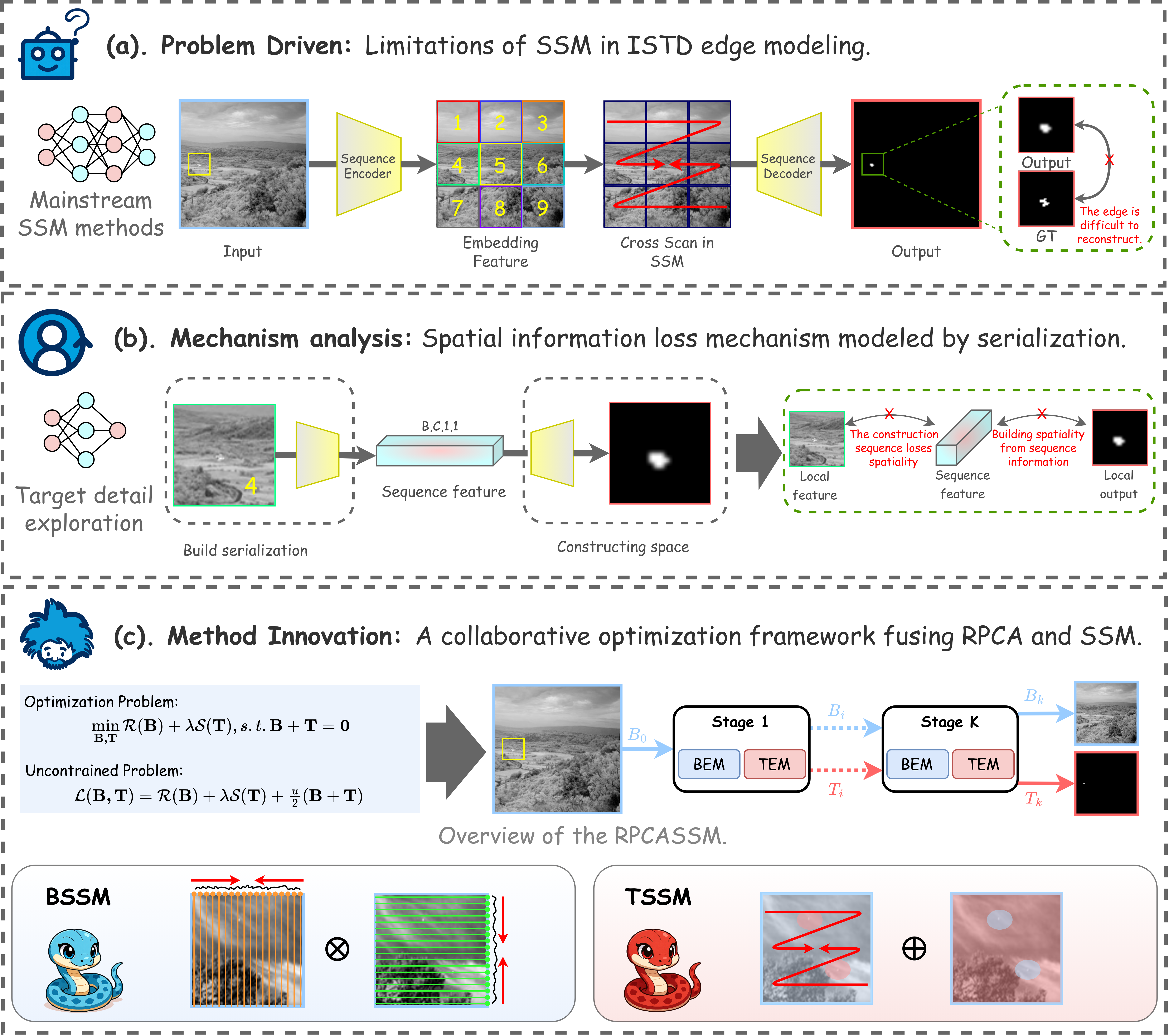}
    \hypertarget{fig:motivate}{}
    \caption{Motivation of RPCASSM framework. (a) Problem-driven: mainstream SSMs are ineffective at modeling edge information for ISTD tasks; (b) Mechanism analysis: sequential construction results in spatial information loss; (c) Methodology: a synergistic optimization framework integrating RPCA and SSM, which achieves background-target decomposition via BSSM/TSSM modules.}
    \label{fig:motivate}
\end{figure}

\IEEEPARstart{I}{nfrared} small target detection (ISTD) is the pivotal technology of infrared search and tracking systems, which holds significant application value in surveillance, security \cite{9714770} and maritime rescue \cite{5730289}.
Infrared small targets are mainly located by infrared sensors in long distance and low visibility conditions, which makes the target signal weak and susceptible to background clutter, resulting in severe challenges in their detection \cite{yuan2025ascnet}.

The shift from model-driven to data-driven methodologies marks a pivotal advancement in ISTD, unlocking new insights and capabilities.
Early model-driven approaches, including filtering \cite{deshpande1999max, rivest1996detection}, local contrast \cite{chen2013local, han2019local}, and low-rank sparse decomposition \cite{gao2013infrared, zhu2019infrared, li2022dense}, were limited by their reliance on manual parameter tuning, leading to poor generalization. 
In contrast, data-driven methods have become the mainstream by automatically learning discriminative features from data, offering superior robustness and adaptability \cite{dai2021asymmetric, dai2021attentional}. 
Recent studies  \cite{yuan2024sctransnet, lu2025lsdssms, chen2024mim} further highlight that modeling long-range dependencies is crucial for capturing global context and boosting detection performance.

In response to this demand, the state space model (SSM) \cite{gu2024mamba, dao2024transformers} is a dynamic system modeling method that has quickly gained attention in the vision community \cite{liu2024vmamba, zhang2025cwnet} due to its ability to effectively model long-distance sequences with linear computational complexity. 
As shown in Fig. \hyperlink{fig:motivate}{1(a)}, visual state-space models (VSSM) based on mainstream paradigms have been successfully applied to image enhancement, object detection and other fields \cite{zhang2025bsmamba, zhang2025irmamba}. 
However, the direct application of mainstream VSSM to infrared small target detection (ISTD) faces fundamental architectural mismatches that stem from the unique characteristics of this domain. 
As shown in Fig. \hyperlink{fig:motivate}{1(b)}, the primary challenge lies in the serialization strategies employed by existing VSSM. 
Mainstream approaches, such as VMamba \cite{gu2024mamba}, typically operate on downsampled feature maps obtained through convolutional stems, which is computationally efficient for general vision tasks. 
However, infrared small targets often occupy merely a few pixels in the original image, making them extremely vulnerable to such early-stage spatial reduction. 
This aggressive downsampling inevitably causes irreversible information loss small targets are either completely eliminated or their critical pixel-level contrast patterns are averaged out with surrounding background pixels. 
While the linear complexity of Mamba theoretically permits processing higher-resolution sequences, the quadratic growth in sequence length when maintaining fine-grained spatial resolution (e.g., stride=1) still poses significant computational challenges. 
Furthermore, the scanning patterns and granularity in current VSSM are optimized for semantic and structural modeling rather than preserving pixel-level extrema crucial for ISTD. 
This is particularly problematic given the low signal-to-noise ratio (SNR) characteristic of infrared imagery, where target saliency depends on subtle local intensity variations rather than semantic features. 

Existing VSSM in the infrared domain \cite{chen2024mim, zhang2025irmamba, li2025smile} have largely adopted these mainstream structural paradigms without addressing the fundamental mismatch between coarse-grained serialization and the fine-grained spatial fidelity required for accurate target edge reconstruction. 
This architectural incompatibility manifests most critically in the difficulty of reconstructing precise target boundaries from sequence features. 
The causality constraints inherent in sequential processing, combined with inadequate spatial resolution preservation, result in blurred or distorted target contours—a fatal flaw for ISTD where accurate localization of minute objects is paramount. 
Consequently, while high-resolution serialization might seem a straightforward remedy, its prohibitive computational cost renders it impractical for real-world deployment. 
This dilemma motivates our search for a novel approach that fundamentally rethinks the serialization strategy to achieve an optimal balance between spatial fidelity preservation and computational efficiency in ISTD. 

Based on the above analysis, as illustrated Fig. \hyperlink{fig:motivate}{1(c)}, we introduce a spatial-domain decomposition prior to resolve the structural mismatch between VSSM and ISTD. 
Specifically, we employ robust principal component analysis (RPCA) \cite{wu2024rpcanet} to decouple an input image X into a low-rank background B and a sparse target component T. 
This decomposition exploits the heterogeneous statistics of ISTD: 
backgrounds are spatially smooth and redundant, whereas targets are sparse, high-contrast extrema. 
By utilizing the heterogeneous information between target regions and background regions, we regard the detection of target regions as that of abnormal signals, thereby constructing an effective spatial probe scanning mechanism(SPCM). 
In the target state space module(TSSM), drawing on the insights from target sparsity and local highlight characteristics, we design deformable target region prompts to cluster the entire spatial domain into target domains and background domains. 
For these two domains, we employ a collaborative feedback factor to control the generation of feedforward matrices for both domains, so as to avoid data imbalance, and construct a deformable prompt scanning mechanism(DPCM) specifically within the target domain. 
Based on the aforementioned designs, we effectively model the edges of infrared small targets. 

Overall, our contributions are as follows:
\begin{itemize}
    \item We propose RPCASSM, an RPCA guided dual branch state space framework that decomposes images into background and sparse target components to enable region adaptive serialization aligned with ISTD’s spatial characteristics.
    \item We design a Spatial Probe Scanning Mechanism (SPCM) to select native resolution regions of interest, a Target SSM (TSSM) with a Deformable Prompt Scanning Mechanism (DPCM) for topology preserving, edge aware modeling, and a Background SSM (BSSM) with coarse, efficient scanning. A collaborative feedback mechanism coordinates both branches and mitigates data imbalance.
    \item We conduct extensive experiments on two benchmark datasets NUDT-SIRST and IRSTD-1K, demonstrating that our RPCASSM outperforms state-of-the-art methods in ISTD.
\end{itemize}

\begin{figure*}[t]
    \centering 
    \includegraphics[scale=0.3]{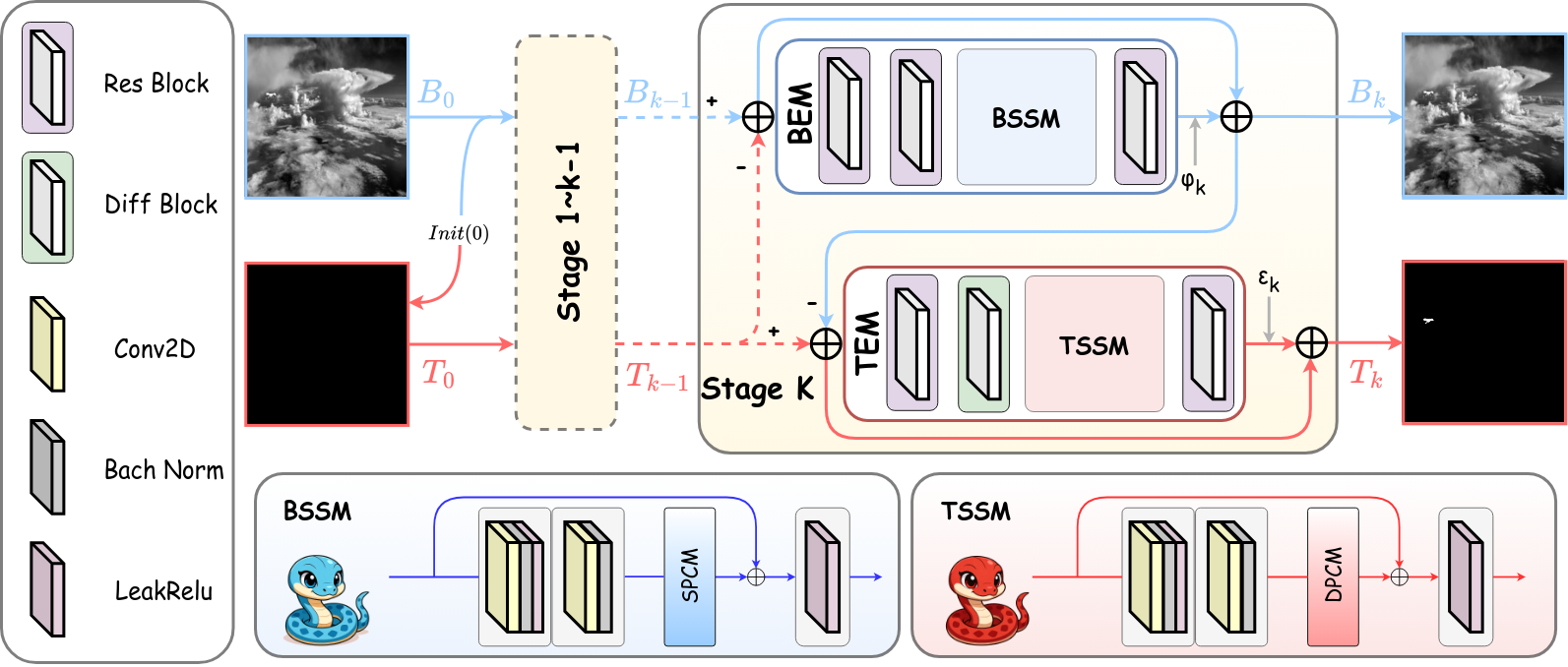}  
    \caption{The overall structure of the RPCASSM. The network is composed of $k$ identical structures, and each layer includes DifBlock, TSSM and BSSM.}
    \label{fig:PCASSM}
\end{figure*}

\section{RELATED WORK}
\label{sec:relatedwork}
\subsection{Infrared Small Target Detection}
Early research on Infrared small target detection (ISTD) mainly follows a model-driven paradigm and highly relies on human-designed prior knowledge \cite{deshpande1999max, chen2013local, gao2013infrared}. These methods are usually based on the physical imaging properties of small objects (such as local gray contrast, shape prior, and the low rank of the background and the sparsity of the target), and construct manually optimized algorithms to achieve detection. However, these traditional methods \cite{han2019local, zhu2019infrared, li2022dense} have significant limitations. On the one hand, their performance depends heavily on the setting of a large number of hyperparameters, which often require manual tuning for different scenes and lack of adaptability. On the other hand, due to the limited semantic representation ability, it is difficult to capture high-level semantic information, which leads to a significant decline in detection robustness in challenging scenes such as low signal-to-noise ratio, complex background clutter and variable target shape.

With the vigorous development of deep learning technology, data-driven methods have been successfully applied to the ISTD field \cite{ni2025point, ying2023mapping, 8734113, 11373245}, promoting the development of the field. Instead of hand-engineered priors, these methods learn discriminative features through data-driven learning. Current research mainly focuses on more effective feature representation from two aspects. In terms of model structure design, researchers are devoted to exploring network architectures more suitable for infrared small target characteristics \cite{zhang2024irsam, huang2025text, pang2025rethinking}. It includes multi-scale feature fusion, attention mechanism embedding and lightweight network design. In terms of loss function constraints, a more refined optimization objective is designed according to the characteristics of infrared small targets \cite{wang2019miss, zhang2022isnet, liu2024infrared, yang2025pinwheel}, Such as suppressing background interference, enhancing target response, and solving the extreme foreground-background imbalance problem. Specifically, some methods make full use of the powerful local feature extraction ability of CNN, and enhance the sensitivity to weak and small targets by designing asymmetric convolution, dilated convolution and other strategies \cite{dai2021asymmetric, li2022dense}. Other methods introduce the Transformer architecture and use the self-attention mechanism to capture the global context and obtain a more effective semantic representation \cite{yuan2024sctransnet, wu2023mtu}, so as to make up for the shortcomings of CNN in long-range dependency modeling. In addition, recent studies have attempted to introduce the State Space Model (SSM) into ISTD \cite{lu2025lsdssms, zhang2025irmamba}.


\subsection{RPCA architecture for ISTD}
RPCANet\cite{wu2024rpcanet} pioneered the fusion of model-driven interpretability with data-driven learning capabilities, providing a new paradigm for infrared small object detection (ISTD).
The method is based on the physical imaging characteristics of infrared images, that is, the background regions are highly correlated (low-rank), and the small targets are spatially isolated (sparsity). Accordingly, RPCANet decomposes the infrared image D into a low-rank background component B and a sparse target component T, and constructs the following optimization model under the framework of robust principal Component Analysis (RPCA): 
\begin{equation}
    \min_{\mathbf{B,T}}\mathcal{L}(\mathbf{B,T}) = \mathcal{R}(\mathbf{B}) + \lambda \mathcal{S}(\mathbf{T}) + \frac{\mu}{2}\|\mathbf{D - B - T}\|_F^2
    \label{eq:rpca}
\end{equation}
where $\mathcal{R}(\cdot)$ and $\mathcal{S}(\cdot)$ are the regularization terms for the low-rank background and sparse target, respectively. 
The parameters $\lambda$ and $\mu$ balance the contributions of these terms. 
This objective is achieved by alternately updating B and T through iterative minimization. 

Inspired by RPCANet, subsequent studies have further explored the potential of this framework and improved it for its limitations. 
Aiming at the problem that the fixed hyperparameters in the original RPCA framework are difficult to adapt to complex and variable scenes, DRPCA-Net\cite{xiong2025drpca} introduces a dynamic parameter generation mechanism and residual grouping strategy, which can adaptively adjust the constraint strength in the decomposition process according to the local characteristics of the input image. 
It effectively alleviates the over-suppression or under-suppression phenomenon of traditional methods in strong clutter background. 
In addition, to overcome the limitations of convolutional operations in capturing long-distance dependencies, LSDSSMs\cite{lu2025lsdssms} creatively combines the low-rank sparse decomposition theory with the emerging State Space Models (SSMs) to construct a unidirectional state space architecture. 
The linear complexity of SSM is used to efficiently model the global context information, which not only inherits the ability of RPCA to suppress background clutter, but also greatly improves the model's ability to capture sequence features of weak targets, achieving a new breakthrough in detection performance. 
Together, these works show that embedding the Inductive Bias of traditional optimization theory into deep learning architectures is an effective way to improve the robustness and accuracy of ISTD tasks.

\subsection{Visual State Space Model for ISTD}

State Space Models (SSM) \cite{gu2024mamba} has risen rapidly in the field of computer vision in recent years due to its excellent ability in modeling long sequences and linear computational complexity. A powerful alternative to the Transformer architecture \cite{vaswani2017attention}. Different from the quadratic complexity of the traditional self-attention mechanism, SSM achieves efficient information transmission through recursive state update. The core of SSM is to capture the dynamic dependencies in sequence data through the recursive evolution of hidden states. Its fundamental equations are as follows:

\begin{equation}
\begin{aligned}
\mathbf{h}_{t} &= \mathbf{A}\mathbf{h}_{t-1} + \mathbf{B}\mathbf{x}_{t} \\
\mathbf{y}_{t} &= \mathbf{C}\mathbf{h}_{t} + \mathbf{D}\mathbf{x}_{t}
\end{aligned}
\end{equation}
where $\mathbf{h}_{t}$ represents the hidden state at time step t, $\mathbf{x}_{t}$ is the input, and $\mathbf{y}_{t}$ is the output. 
The matrices $\mathbf{A}$, $\mathbf{B}$, $\mathbf{C}$, and $\mathbf{D}$ represent the state transition matrix, input matrix, output matrix, and feedthrough matrix, respectively.

Inspired by this, the field of infrared small target detection (ISTD) has begun to actively explore the application potential of SSM, trying to combine its powerful sequence modeling ability with infrared imaging characteristics. MiM-ISTD\cite{chen2024mim} achieves the collaborative unification of global context awareness and local detail focus by constructing the internal and external dual state space architecture. SMILE\cite{li2025smile} combines the spatial domain and frequency domain modeling, and uses the expression advantage of SSM in frequency domain sequence to enhance the suppression ability of periodic background clutter. IRMamba\cite{zhang2025irmamba} improves the discrimination between weak targets and complex backgrounds by introducing local pixel difference features. Although the above works have made significant progress, they essentially follow the mainstream visual state Space Model (VSSM) paradigm designed for natural images, which fails to fully adapt to the special needs of infrared small target detection: On the one hand, infrared images generally have problems such as low signal-to-noise ratio, missing target texture and weak features, and the existing SSM architectures lack robust design for such low-quality inputs. On the other hand, the infrared small target presents highly sparse distribution in space and has continuous motion characteristics in time series. The state space design of the current method is relatively fixed, which is difficult to flexibly capture the spatial sparsity and temporal dynamics of the target. Therefore, designing a state space modeling method that is more suitable for the characteristics of ISTD tasks is still a key direction in this field. 

\section{METHODOLOGY}
\label{sec:method}
In this part, we introduce the structure design of RPCASSM, and then focus on the design of background state space module and target state space module.

\begin{figure}[t]
    \centering 
    \includegraphics[scale=0.18]{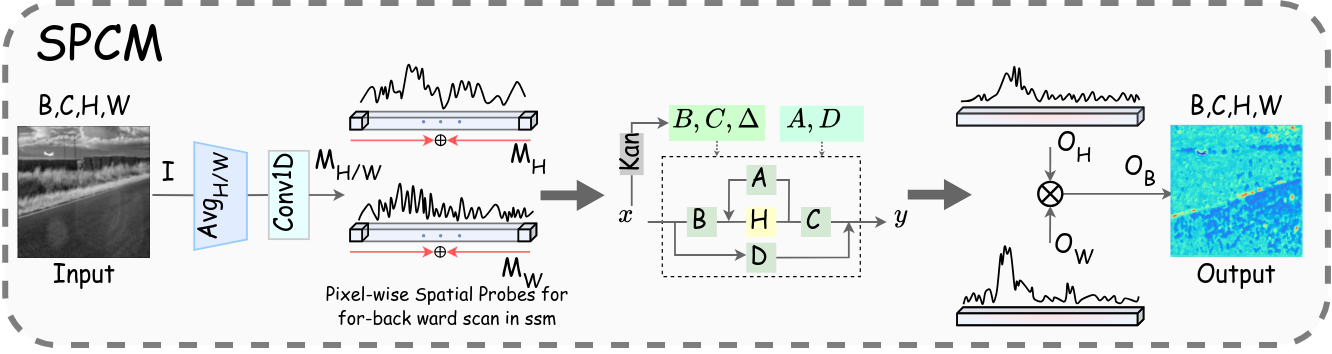}  
    \caption{The structure of the SPCM in the BSSM. }
    \label{fig:SPCM}
\end{figure}

\subsection{Overall Framework}
As shown in \cref{fig:PCASSM}, the proposed RPCASSM architecture aims to address the ISTD challenge by constructing accurate state-space modeling through the spatial nature of ISTD, thereby avoiding the limitations brought by the mainstream structural paradigm. 
As shown in \cref{eq:rpca}, based on the physical modeling of the infrared image by RPCA, we construct the unconstrained target of the infrared image as follows: 
\begin{equation}
    \min_{\mathbf{B,T}}\mathcal{L}(\mathbf{B,T}) = \mathcal{R}(\mathbf{B}) + \lambda \mathcal{S}(\mathbf{T}) + \frac{\mu}{2}\|\mathbf{B + T}\|_F^2
\end{equation}
Different from the conventional RPCA modeling mentioned above, we regard $\mathbf{B,T}\in \mathbb{R}^{H\times W}$ as the background and object mask modeling in the iterative process, which makes them opposite. The optimization is carried out by the iterative alternating optimization method, and the mathematical formula is as follows:
\begin{equation}
\begin{aligned}
\mathbf{B}^k&=\mathbf{B}^{k-1} - \mathbf{T}^{k-1} + \varphi^{k}\mathcal{R}^{k}(\mathbf{B}^{k-1} - \mathbf{T}^{k-1}) \\
\mathbf{T}^k&=\mathbf{T}^{k-1} - \mathbf{B}^{k} + \psi^{k}\mathcal{S}^{k}(\mathbf{T}^{k-1} - \mathbf{B}^{k})
\end{aligned}
\end{equation}
here, the upscript $k$ represents the sequence number of the stage, $\mathbf{B}^k$ and $\mathbf{T}^k$ represent the background and target at stage k respectively, $\varphi^{k}$ and $\psi^{k}$ are learnable parameters, and $\mathcal{R}^{k}(\cdot)$ and $\mathcal{S}^{k}(\cdot)$ are the background extraction module(BEM) and target extraction module(TEM) respectively.

In the BEM, background information is further extracted iteratively from $\mathbf{B}_{k-1}$ and $\mathbf{T}_{k-1}$ to ensure low-rank background modeling, which is mainly composed of Resblock \cite{he2016deep} and BSSM(\cref{sec:BSSM}). 
In the TEM, more effective sparse targets are modeled, and more accurate target information is iteratively extracted from $\mathbf{B}_{k}$ and $\mathbf{T}_{k-1}$ feedback from BEM, mainly composed of Resblock, Difblock and TSSM(\cref{sec:TSSM}). 
The design of the Difblock performs convolution operation through the spatial differential feature, which can roughly emphasize the saliency of the target feature, and is specifically described as follows.
\begin{equation}
\begin{aligned}
\phi(\cdot) &= \sum_{i=1}^{N}\sum_{j=1}^{N}w_{ij} \cdot (N^{2}f_{ij} - \sum_{k=-\frac{N}{2}}^{\frac{N}{2}}\sum_{l=-\frac{N}{2}}^{\frac{N}{2}}f_{i+k,j+l}) \\
&= \sum_{i=1}^{N}\sum_{j=1}^{N}(N^{2}w_{ij} - \sum_{k=-\frac{N}{2}}^{\frac{N}{2}}\sum_{l=-\frac{N}{2}}^{\frac{N}{2}}w_{i+k,j+l}) \cdot f_{ij}
\end{aligned}
\end{equation}
here, $\phi(\cdot)$ represents the difference operation, $f_{ij}$ and $w_{ij}$ represent the pixel value and weight value at position (i,j) within the sliding window respectively, and $N$ represents the size of the convolution kernel.

In order to effectively enhance the gradient propagation and feature representation ability, we adopt the way of progressive loss constraint.
RPCASSM will generate sparse targets $\mathbf{T}^i \in [1, k]$ at each stage, and we regard $\mathbf{T}^{k}$ as the final prediction result. The loss function $\mathcal{L}_{IoU}$ is then used for progressive constraints, as shown in the following details:
\begin{equation}
\begin{aligned}
\mathcal{L}_{IoU}(\mathbf{T}, \mathbf{Y}) & = 1- \frac{\mathbf{T} \cap \mathbf{Y}}{\mathbf{T} \cup \mathbf{Y}} \\ 
&= 1- \frac{\sum_{n}(\mathrm{T}_n \cdot \mathrm{Y}_n)}{\sum_{n}(\mathrm{T}_n + \mathrm{Y}_n - \mathrm{T}_n \cdot \mathrm{Y}_n)}
\end{aligned}
\end{equation}
here, $\mathbf{Y}$ represents the ground truth of the target, and $\mathrm{T}_n$ and $\mathrm{Y}_n$ represent the pixel values at position n of $\mathbf{T}$ and $\mathbf{Y}$ respectively. The overall loss function is defined as follows:
\begin{equation}
\mathcal{L}_{total} = \mathcal{L}_{IoU}(\mathbf{T}^{k}, \mathbf{Y}) + \sum_{i=1}^{k-1}\alpha \mathcal{L}_{IoU}(\mathbf{T}^{i}, \mathbf{Y})
\end{equation}
Here $\alpha$ is the weight parameter used to control the expression of intermediate sparse target capability.

\subsection{Background State Space Module}
\label{sec:BSSM}
In order to overcome the problem that the existing VSSM are easily limited by local spatial information when constructing long-distance dependencies, this paper proposes a Background State Space Module (BSSM). 
The proposed module is based on the heterogeneous characteristics of saliency between infrared small target and background. 
By effectively segmenting heterogeneous information on the spatial axis, the refined modeling of state space relations and the enhancement of long-distance dependence are realized.

As shown in \cref{fig:SPCM}, the core design of the proposed method is the introduction of Spatial Probe Scanning Mechanism (SPCM), which systematically captures heterogeneous signals in the image through bidirectional state space modeling on the spatial axis, so as to realize the global modeling of background information. Firstly, the spatial axis representation is obtained, which is formulated as follows:
\begin{equation}
    \mathbf{M}_{H/W} = Conv1d(Avg_{H/W}(\mathbf{I}))
\end{equation}
$\mathbf{I} \in \mathbb{R}^{C\times H \times W}$ represents the input background features and $\mathbf{M}_{H/W} \in \mathbb{R}^{C \times H/W}$ represents the representation of the extracted spatial axes. 
Then we build a SSM to find heterogeneous information, as shown in the following formula:
\begin{equation}
    B,C,\Delta = Kan(\mathbf{M}_{H/W})
\end{equation}
\begin{equation}
    \mathbf{O}_{H/W} = \mathbb{S}_{B,C,\Delta}^{for}(\mathbf{M}_{H/W}) + \mathbb{S}_{B,C,\Delta}^{back}(\mathbf{M}_{H/W})
\end{equation}
here, we use $Kan$\cite{liu2024kan} to generate parameters $B,C,\Delta$. $\mathbb{S}_{B,C,\Delta}^{for / back}(\cdot)$ denotes the SSM parameterized using $B,C,\Delta$, where $for$ and $back$ denotes forward scan and backward scan, respectively. 
The feature $\mathbf{O}_{H/W}$ is generated using $\mathbf{M}_{H/W}$ through $\mathbb{S}_{B,C,\Delta}^{for / back}(\cdot)$. 
The background features are then constructed using the following formula:
\begin{equation}
    \mathbf{O}_{H},\mathbf{O}_{W} = \mathbf{O}_{H/W}
\end{equation}
\begin{equation}
    \mathbf{O}_{B} = \mathbf{O}_{H} \cdot \mathbf{O}_{W}
\end{equation}
where, $\mathbf{O}_{H}, \mathbf{O}_{W}\in \mathbb{R}^{C\times H/W}$ represent the features of the axis, and $\mathbf{O}_{B} \in \mathbb{R}^{C\times H \times W}$ represents the background features after construction.

\subsection{Target State Space Module}
\label{sec:TSSM}
For further accurate object edge modeling, sparse object edge joint state space modeling is preserved. 
In this paper, we propose the Target State Space Module (TSSM).
The proposed module is based on the sparsity of infrared small target and the local highlight property of infrared target. 
By effectively modeling the local highlight dynamic region in the spatial domain, the module can focus on a small part of the target area to preserve the fine edge of the target and avoid the demand for computing resources brought by the same resolution. 
\begin{figure}[t]
    \centering 
    \includegraphics[scale=0.038]{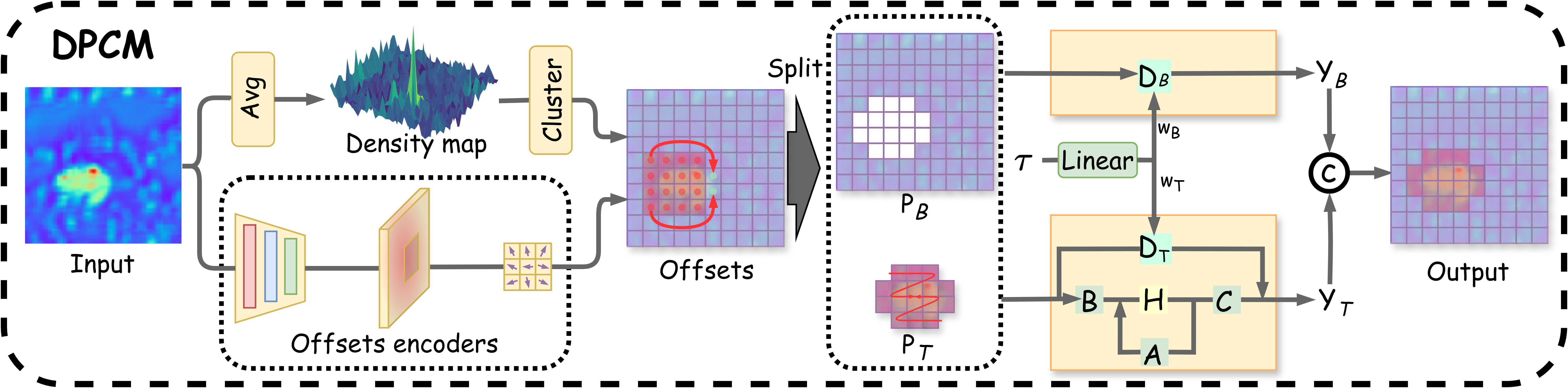}  
    \caption{The structure of the DPCM in the TSSM}
    \label{fig:DPCM}
\end{figure}

As shown in \cref{fig:DPCM}, the core design of our method is Deformable prompt scanning mechanism(DPCM). 
This mechanism obtains the prompt area features by clustering the target area through density features in the spatial domain, and learns the deformable interpolation using the input features. 
Then use this as a reference to adjust the deformable prompt area. 
Then, the long-range dependent SSM is established for the target region, and the output of the background region is mapped by the feedforward matrix, and the two are associated by the collaborative control matrix. 

The first step of DPCM is to extract and cluster density maps in the spatial domain, and learn deformable interpolation and segmentation into target cue regions and background regions, which are detailed as follows:
\begin{equation}
    \mathbf{P}_{T}, \mathbf{P}_{B} = Split(G_{cluster}(Avg(\mathbf{I})),G_{offsets}(\mathbf{I}))
\end{equation}
here, $\mathbf{P}_{T}$ and $\mathbf{P}_{B}$ represent target region and background region of the prompt obtained by clustering, respectively. 
$Split(\mathrm{M}_{mask}, \mathrm{M}_{value})$  means to split $\mathrm{M}_{value}$ using the target region mask provided by $\mathrm{M}_{mask}$. 
$G_{offsets}(\cdot)$\cite{dai2017deformable} stands for deformable learning to adjust offset indirectly. 
$Avg(\cdot)$ stands for average pooling layer. 
$G_{cluster}$ represents the spatial clustering of target and background regions using the highlighting property. 
The $G_{cluster}$ of algorithm is as follows:
\begin{equation}
    \mathrm{M}_{mask} = \mathbbm{1} \left( \mathrm{M}_{density} > T_s \right)
\end{equation}
here our binary mask $\mathrm{M}_{mask}$ applies an adjustable threshold $T_s$ through the density heatmap $\mathrm{M}_{density}$.
Where at least $\gamma$ steps of adjustment are required for $T_s$ to generate at least one foreground region:
\begin{equation}
\gamma = \min \left[\gamma \in \mathbb{N} | \mathbbm{1} (\mathrm{M}_{density} > T_s) > 0 \right],
\end{equation}
\begin{equation}
T_s = T_{init} - \gamma \cdot \Delta T
\end{equation}
Where $T_{init}$ refers to the initial threshold and $\Delta T$ is the decreasing quantity.

Then the obtained $\mathbf{P}_T$ and $\mathbf{P}_B$ construct a sparse SSM for the target region $\mathbf{P}_T$, $\mathbf{P}_B$ performs feedforward mapping, and the two are associated through the cooperative control matrix $\tau$ to avoid data imbalance. 
The detailed formula is expressed as follows:
\begin{equation}
    \mathbf{D}_B = Linear_{w_B}(\tau), \mathbf{D}_T = Linear_{w_T}(\tau)
\end{equation}
The above expression not only preserves the relationship between the two, but also preserves their independence. 
Where $\mathbf{D}_T$ and $\mathbf{D}_B$ represent the feedforward matrix of the target and background, respectively.
\begin{equation}
    \mathbf{Y}_B =  \mathbf{D}_B \cdot \mathbf{P}_B
\end{equation}
\begin{equation}
    \mathbf{Y}_T = \mathbb{S}_{\mathbf{D}_T}^{CSM}(\mathbf{P}_T)
\end{equation}
\begin{equation}
    \mathbf{M}_{O} = Cat(\mathbf{Y}_{B}, \mathbf{Y}_{T}) 
\end{equation}
Above $\mathbb{S}_{\mathbf{D}_T}^{CSM}$ denotes the SSM using the parameterized feedforward matrix of $\mathbf{D}_T$, and $CSM$ \cite{liu2024vmamba} denotes the cross-scan module. 
Cat represents the concatenation of the generated feature $\mathbf{Y}_B$ and $\mathbf{Y}_T$ over space, and $\mathbf{M}_O$ is the output heatmap.

\begin{table*}[h]
\caption{Quantitative results of different methods. Comparisons with SOTA methods on NUDT-SIRST and IRSTD-1K in Params(M), mIoU(\%), Pd(\%), Fa($10^{-6}$), F-measure(\%). The best is show in bold and the second best is underlined.}
\label{tab:compare}
\centering
\begin{tabular}{l|c|c c c c|c c c c}
\hline
\multirow{2}{*}{Method} & 
\multirow{2}{*}{Params} & 
\multicolumn{4}{c|}{NUDT-SIRST} & 
\multicolumn{4}{c}{IRSTD-1K} \\
\cline{3-10}
 & & mIoU & Pd & Fa & F-measure & mIoU & Pd & Fa & F-measure \\
\hline
ACM(WACV2021)\cite{dai2021asymmetric} & 0.52 & 66.80 & 95.24 & 22.52 & 80.09 & 52.70 & 84.54 & 65.86 & 69.03 \\
ALCNet(TGRS2021)\cite{dai2021attentional} & \underline{0.42} & 71.08 & 95.24 & 19.32 & 83.10 & 57.82 & 89.00 & 52.47 & 73.27 \\
DNANet(TIP2022)\cite{li2022dense} & 4.69 & 92.14 & 97.64 & 9.39 & 95.91 & 62.87 & 91.13 & 55.63 & 77.20 \\
AGPCNet(TAES2023)\cite{zhang2021agpcnet} & 12.36 & 83.72 & 97.14 & 17.67 & 91.14 & 61.41 & 89.69 & 26.62 & 76.09 \\
UIUNet(TIP2023)\cite{wu2022uiu} & 50.54 & 90.52 & \underline{97.80} & 8.34 & 95.02 & 65.69 & \underline{91.24} & 23.47 & 79.31 \\
MSHNet(CVPR2024)\cite{liu2024infrared} & 4.06 & 91.64 & 97.78 & \underline{3.58} & 95.64 & 56.06 & 88.66 & 29.75 & 71.84 \\
RPCANet(WACV2024)\cite{wu2024rpcanet} & 0.67 & 91.30 & 97.46 & 17.99 & 95.45 & 60.53 & 88.66 & 37.80 & 75.41 \\
SCTransNet(TGRS2024)\cite{yuan2024sctransnet} & 11.19 & 94.58 & 97.13 & 3.70 & \underline{97.21} & 66.46 & 90.03 & 27.86 & 79.85 \\
LSDSSMs(TGRS2025)\cite{lu2025lsdssms} & \textbf{0.37} & \underline{94.59} & 97.52 & 3.77 & 97.22 & \underline{68.14} & 86.60 & 24.80 & \underline{81.05} \\
DRPCA-Net(TGRS2025)\cite{xiong2025drpca} & 1.16 & 94.19 & 97.41 & 13.71 & 97.01 & 65.23 & 90.72 & \underline{23.09} & 78.95 \\
RPCASSM(Ours)& 0.45 & \textbf{95.98} & \textbf{98.62} & \textbf{2.22} & \textbf{97.95} & \textbf{68.44} & \textbf{92.44} & \textbf{22.31} & \textbf{81.26} \\
\hline
\end{tabular}
\end{table*}

\begin{table}[t]
\caption{Ablation studies were performed on each module based on the results of mIoU (\%), Pd (\%), Fa($10^{-6}$), F-measure (\%) and Flops(G) on NUDT-SIRST}
\centering
\label{tab:ablation}
\small
\setlength{\tabcolsep}{3.5 pt}
\begin{tabularx}{\linewidth}{@{}ccccccccc@{}}
\toprule\toprule
DifBlock & TSSM & BSSM & mIoU & Pd & Fa & F-measure \\
\midrule 
\XSolidBrush & \XSolidBrush & \XSolidBrush  &  91.60 & 97.41 & 18.49 & 95.61  \\
\CheckmarkBold & \XSolidBrush & \XSolidBrush  & 91.75 & \underline{97.78} & 16.10 & 95.70 \\
\CheckmarkBold & \CheckmarkBold & \XSolidBrush & \underline{93.67} & 97.65 & \underline{5.05} & \underline{96.73}  \\
\CheckmarkBold & \CheckmarkBold & \CheckmarkBold & \textbf{95.98} & \textbf{98.62} & \textbf{2.22} & \textbf{97.95}  \\
\bottomrule
\end{tabularx}
\end{table}

\begin{table}[t]
\caption{Experimental study to explore the impact of hyperparameters on model performance on the results of mIoU (\%),Pd (\%),Fa($10^{-6}$), and F-measure (\%) on NUDT-SIRST}
\label{tab:hyperparameters}
\centering
\small  
\setlength{\tabcolsep}{3.5pt}  
\begin{tabular}{@{}p{2.1cm}p{1.4cm}p{1.4cm}p{1.4cm}p{1.4cm}@{}}  
\toprule\toprule
Hyper-param & mIoU  & Pd  & Fa  & F-measure \\
\midrule
\multicolumn{5}{@{}c@{}}{The number of Stage \( \mathrm{k} \)} \\
\midrule
$k$=1 & 94.09 & 97.67 & 5.49 & 96.96 \\
$k$=2 & 95.01 & 98.20 & 4.62 & 97.44 \\
\textbf{$k$=3} &\textbf{95.98} & \textbf{98.62} & \textbf{2.22} & \textbf{97.95} \\
$k$=4 & 95.42 & 98.52 & 2.94 & 97.66 \\
$k$=5 & \underline{95.85} & \underline{98.61} & \underline{2.71} & \underline{97.80} \\
\midrule
\multicolumn{5}{@{}c@{}}{The weights $\alpha$ of $\mathcal{L}_{IoU}(\cdot,\cdot)$} \\
\midrule
\textbf{$\alpha$=0.01} & \textbf{95.98} & \textbf{98.62} & \textbf{2.22} & \textbf{97.95} \\
$\alpha$=0.03 & 95.33 & 98.33 & 6.66 & 97.61 \\
$\alpha$=0.05 & 95.18 & 98.10 & 3.78 & 97.53 \\
$\alpha$=0.07 & \underline{95.45} & \underline{98.41} & 3.11 & \underline{97.67} \\
$\alpha$=0.10 & 95.02 & 98.21 & \underline{2.88} & 97.45 \\
\midrule
\multicolumn{5}{@{}c@{}}{The value of $T_{init}$} \\
\midrule
$T_{init}$=0.1 & \underline{95.70} & 98.10 & \textbf{2.13} & \underline{97.80} \\
$T_{init}$=0.3 & 95.46 & \underline{98.54} & 3.36 & 97.68 \\
$T_{init}$=0.5 & 95.39 & 98.46 & 4.64 & 97.64 \\
\textbf{$T_{init}$=0.7} & \textbf{95.98} & \textbf{98.62} & \underline{2.22} & \textbf{97.95} \\
$T_{init}$=1.0 & 95.50 & 98.31 & 2.55 & 97.70 \\
\bottomrule
\end{tabular}
\end{table}

\begin{table*}[t]
\centering
\caption{The Auc of the sota method with different thresholds was experimented on the nudt-sirst and irstd-1k datasets.}
\label{tab:ROC}
\resizebox{\textwidth}{!}{
\begin{tabular}{c c c c c c c c c c c c c}
\toprule\toprule
Dataset & Index & ACM & ALCNet & DNANet & AGPCNet & AGPCNet & MSHNet & RPCANet & SCTransNet & DRPCANet & LSDSSMs & RPCASSM \\
\hline
\multirow{2}{*}{IRSTD-1K}
& $\mathrm{AUC_{FPR=0.5}}$ & 0.3875 & 0.4200  & 0.5211 & 0.5127 & 0.4950 & \textbf{0.5683} & 0.4040 & 0.4902 & 0.5030 & 0.4827 & \underline{0.5371} \\
& $\mathrm{AUC_{FPR=1.0}}$ & 0.5247 & 0.5765  & 0.6509 & 0.6386 & 0.6541 & \underline{0.6872} & 0.5838 & 0.6290 & 0.6464 & 0.6288 & \textbf{0.6916} \\
\hline
\multirow{2}{*}{NUDT-SIRST}
& $\mathrm{AUC_{FPR=0.5}}$ & 0.3816 & 0.4039 & 0.8949 & 0.6707 & 0.8146 & 0.9084 & 0.8158 & 0.9078 & 0.8628 & \underline{0.9196} & \textbf{0.9340}\\
& $\mathrm{AUC_{FPR=1.0}}$ & 0.5120 & 0.5546 & 0.9210 & 0.7732 & 0.8751 & 0.9306 & 0.8702 & 0.9324 & 0.9101 & \underline{0.9432} & \textbf{0.9529}\\
\midrule
\end{tabular}
}
\end{table*}

\section{EXPERIMENT}
\label{sec:experiments}

\subsection{Experiment details}

\subsubsection{Evaluation Metrics}
In order to comprehensively compare the performance of the proposed RPCASSM with other state-of-the-art methods, we used five evaluation metrics, including the number of model parameters (M), the mean intersection over Union (mIoU), the probability of detection (Pd), the false alarm rate (Fa), and the F-measure. The number of model parameters (M) is used to measure the complexity of the model. The Mean Intersection over Union (mIoU) was used as the pixel-level evaluation index to calculate the mean intersection over union of all objects. The detection probability (Pd) represents the proportion of correctly detected target pixels over all true target pixels. The false alarm rate (Fa) measures the proportion of false detection pixels in the total pixels of the image. F-measure(F) is based on the harmonic mean of precision and recall, and aims to balance the relationship between accurately detecting true examples and reducing false negatives. The above five indicators are defined as follows:
\begin{equation}
    M =  \sum_{l=1}^{L}{N_l}
\end{equation}

\begin{equation}
    mIoU = \frac{A_i}{A_u} = \frac{\sum_{i=1}^{N}{TP[i]}}{\sum_{i=1}^{N}{(T[i] + P[i] - TP[i])}}
\end{equation}

\begin{equation}
    F = \frac{2 \times Prec \times Rec}{Prec + Rec} 
\end{equation}

\begin{equation}
    Pd = \frac{N_{pred}}{N_{all}}
\end{equation}

\begin{equation}
    Fa = \frac{N_{false}}{P_{total}}
\end{equation}

where, $L$ represents the total number of layers of the network and $N_l$ represents the number of parameters in the $l$ layer. 
In mIoU calculation, $A_i$ and $A_u$ represent the pixel area of the Intersection and Union of the predicted region and the true region, respectively. 
$N$ is the total number of samples; 
$TP[\cdot]$ is the number of True Positive pixels; 
$T[\cdot]$ and $P[\cdot]$ represent the total number of positive class pixels in the Ground Truth label and prediction result, respectively; 
$Prec$ and $Rec$ stand for Precision and Recall, respectively;
$N_{pred}$ denotes the number of correctly predicted targets and $N_{all}$ denotes the total number of targets in the image. 
$N_{false}$ represents the number of false predicted target pixels, and $P_{total}$ represents the total number of image pixels.

\begin{figure*}[t]
    \centering 
    \includegraphics[scale=0.2]{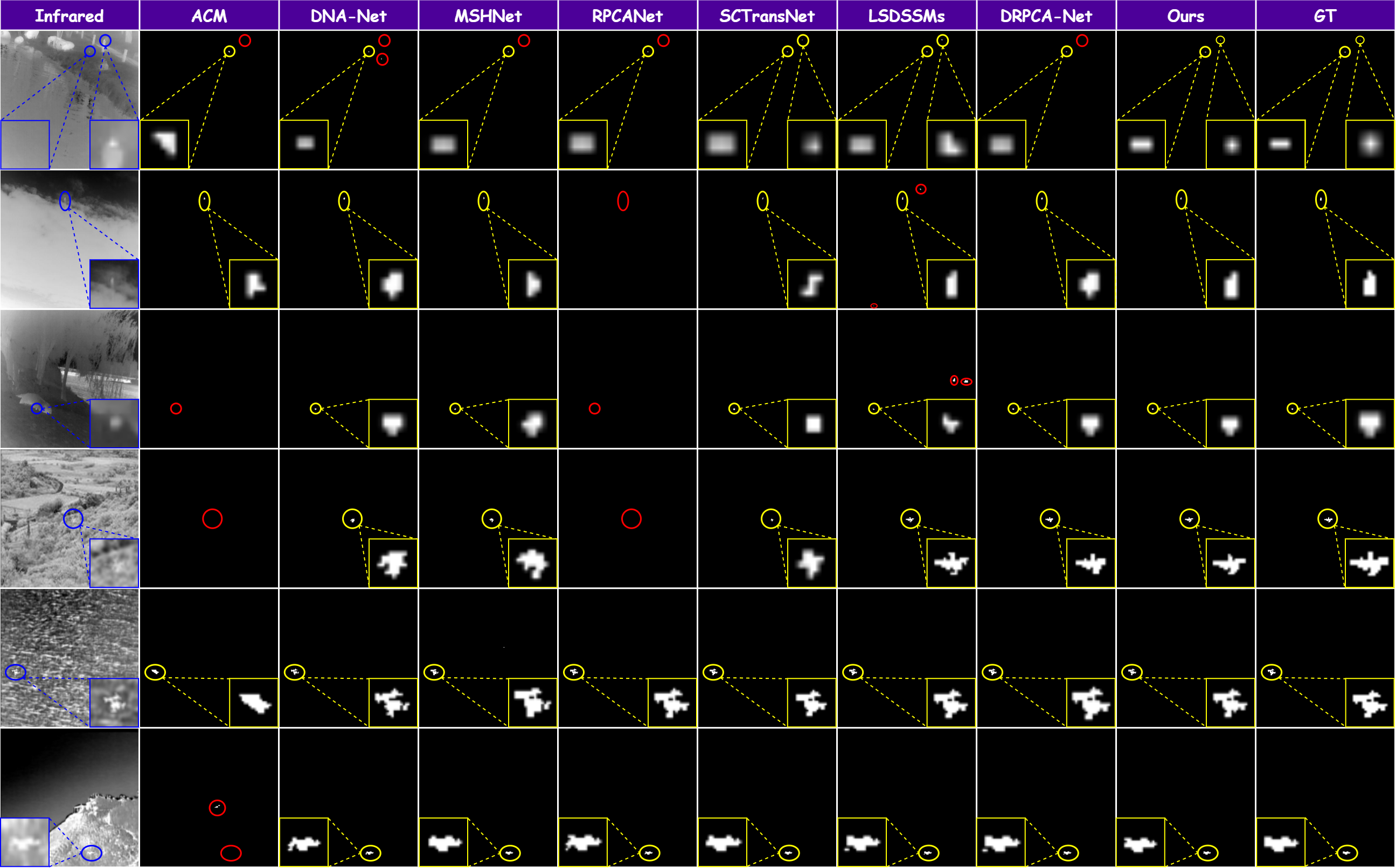}  
    \caption{Visual comparison of detection results on sample images from IRSTD-1K (first three rows) and NUDT-SIRST (last three rows). Yellow and red regions denote correct detections and missed/false alarms, respectively.}
    \label{fig:PCASSM.VISIMG}
\end{figure*}

\begin{figure*}[t]
    \centering 
    \includegraphics[scale=0.2]{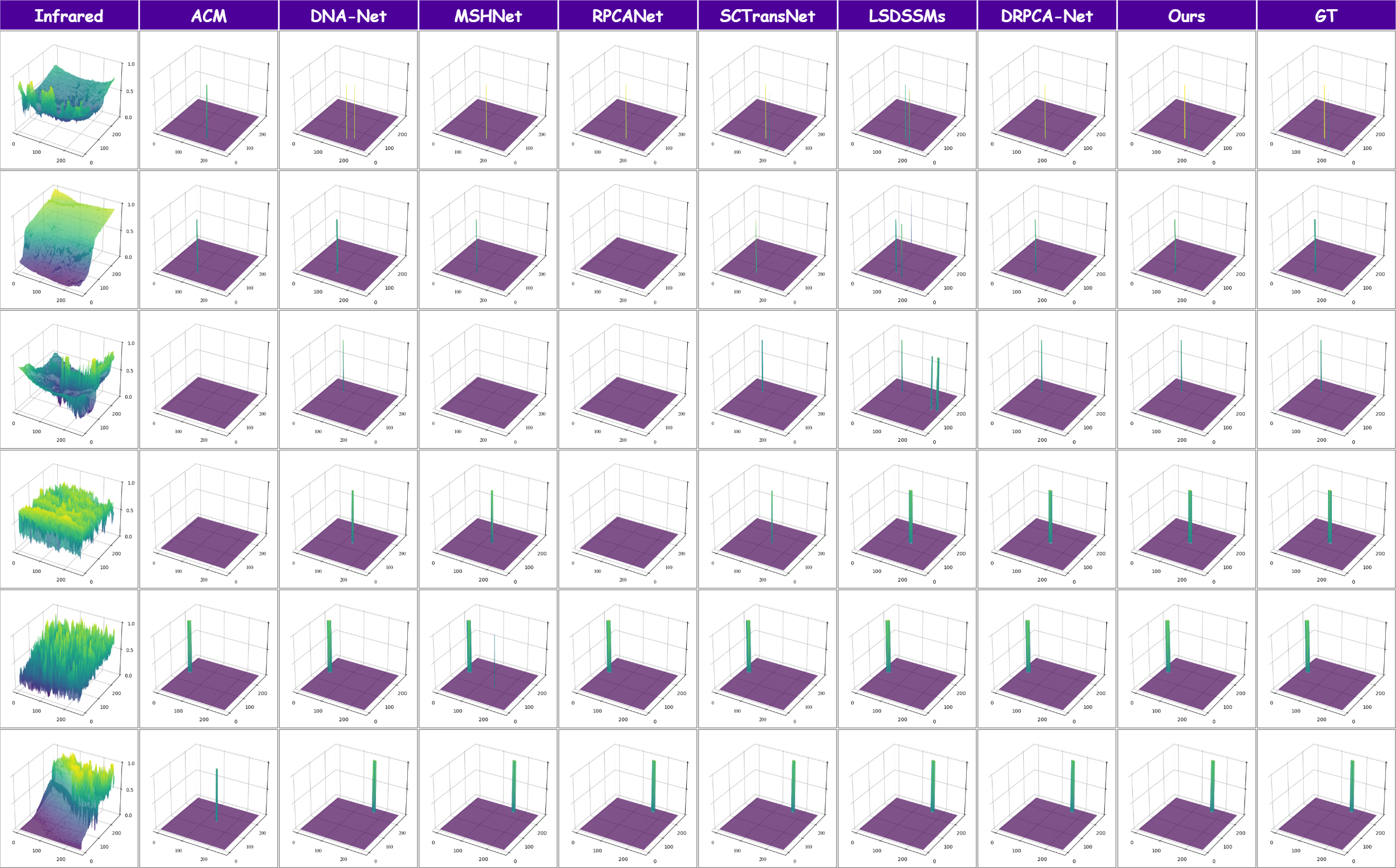}  
    \caption{3D visualizations of the saliency maps produced by different methods on the six test images are presented as the counterparts to \cref{fig:PCASSM.VISIMG}.}
    \label{fig:PCASSM.VIS3D}
\end{figure*}

\begin{figure}[t]
  \centering
  \subfloat[ROC on NUDT-SIRST.]{\includegraphics[width=0.48\linewidth]{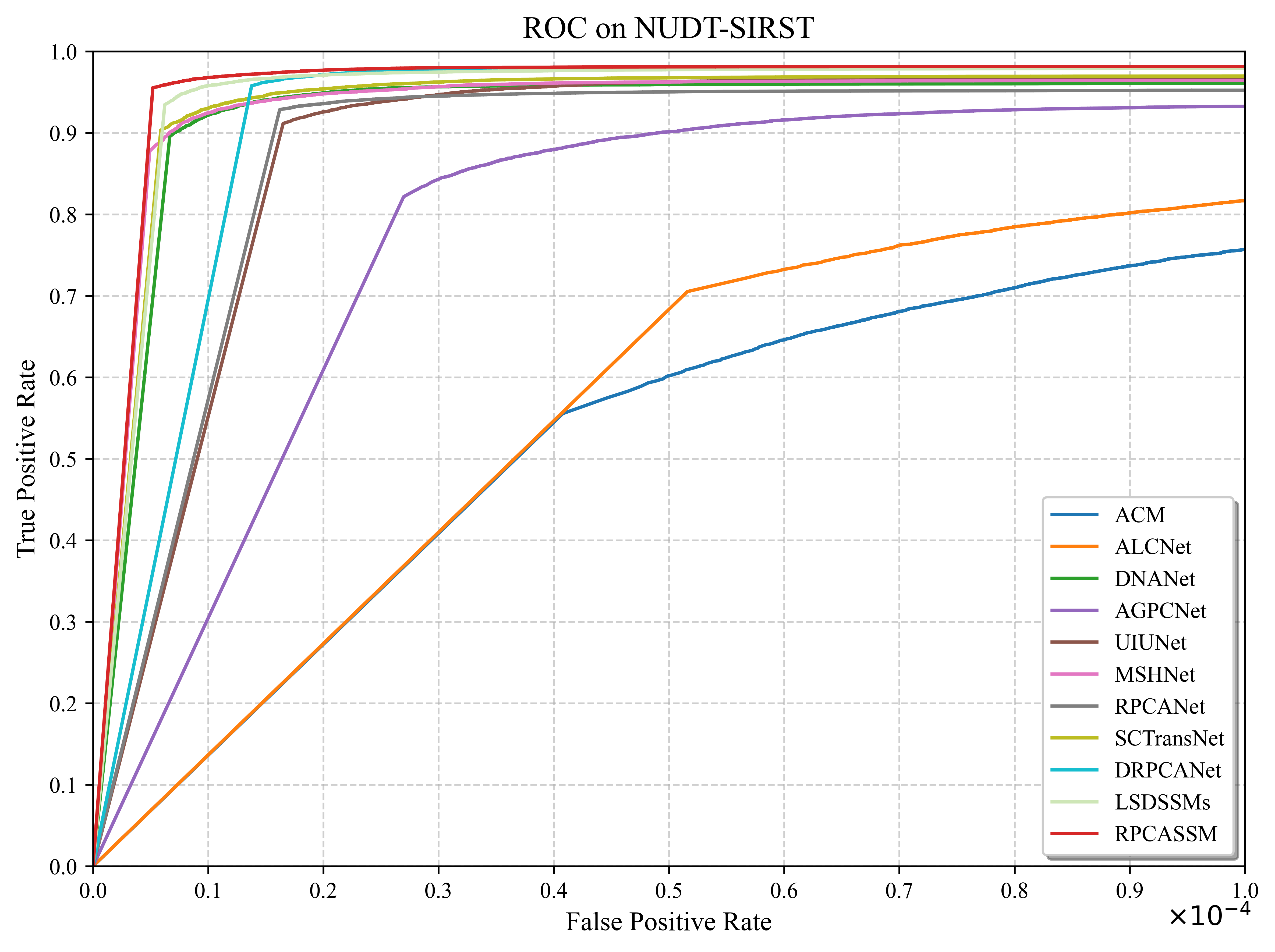}\label{fig:ROC-a}}
  \hfill
  \subfloat[ROC on IRSTD-1K.]{\includegraphics[width=0.48\linewidth]{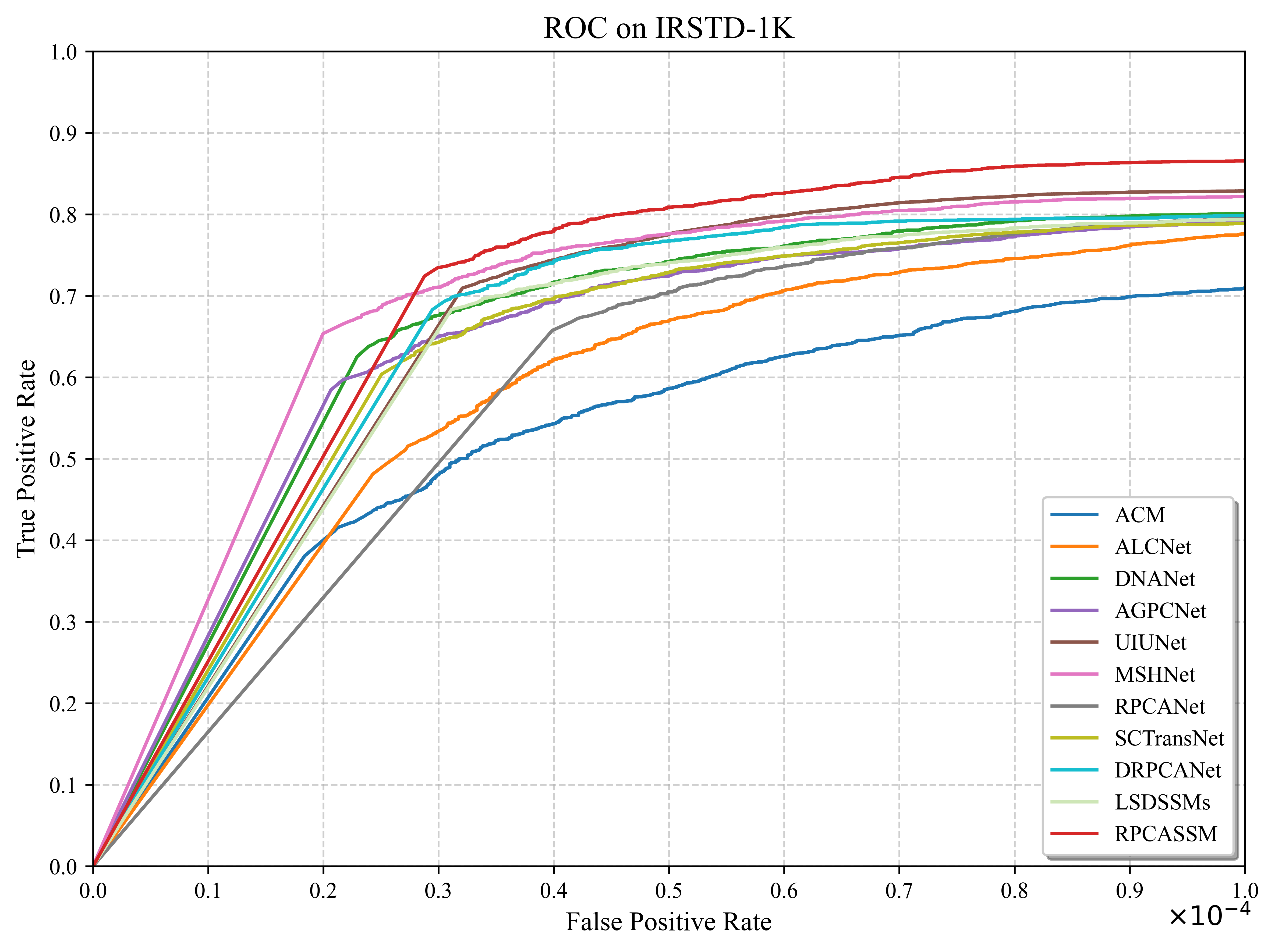}\label{fig:ROC-b}}
  \caption{ROC curves of different methods in NUDT-SIRST and IRSTD-1K.}
  \label{fig:ROC}
\end{figure}

\begin{figure*}[t]
    \centering 
    \includegraphics[scale=0.066]{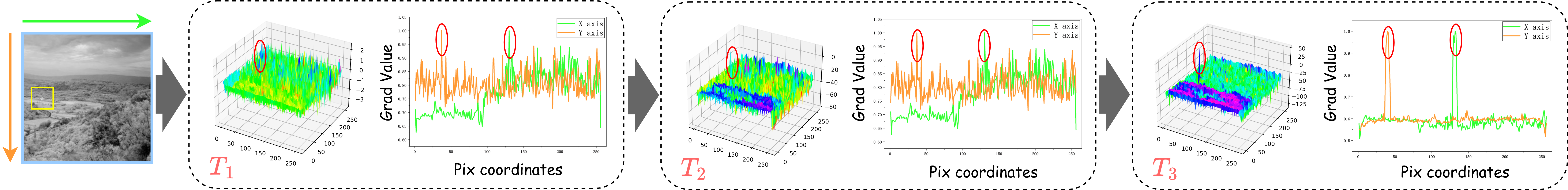}  
    \caption{Stage-wise target heatmaps of the RPCASSM model, demonstrating the heterogeneous signal characteristics of the target along the single-axis direction.}
    \label{fig:TARGET}
\end{figure*}

\subsubsection{Datasets}
We comprehensively validate RPCASSM on two benchmark datasets, NUDT-SIRST\cite{li2022dense} and IRSTD-1K\cite{zhang2022isnet}. These datasets are famous for their rich scene diversity, including typical difficult points such as multi-target interference, weak point sources and low signal-to-noise ratio. In order to eliminate the data distribution bias and ensure the fairness of the evaluation, we use a 1:1 random splitting scheme to construct the training and testing set. This setting not only makes full use of high-quality labeled data, but also forces the model to be tested under extremely complex visual conditions, which strongly proves the effectiveness and superiority of the proposed method.

\subsubsection{Training Details}
All experiments were conducted using the PyTorch deep learning framework on a single NVIDIA GeForce RTX 4070 Ti SUPER GPU. 
To balance GPU memory consumption with the stability of gradient updates, we set the batch size to 8 and trained the model in an end-to-end manner for 800 epochs.
Regarding the optimization strategy, we employed the Adam optimizer with an initial learning rate of $5\times10^{-4}$.
To prevent overfitting and accelerate convergence, a learning rate decay schedule was implemented.
Specifically, the key trainable parameters, $\varphi_k$ and $\varepsilon_k$, within our network were initialized to 0.1 to ensure the model starts in a stable search space during the early training stages.
For data preparation, we utilized two challenging public benchmark datasets: NUDT-SIRST and IRSTD-1K. To eliminate the impact of scale variations on model performance, all input images were uniformly resampled to a resolution of $256\times256$ via bilinear interpolation and subjected to standard normalization. Furthermore, data augmentation techniques, including random flipping and rotation, were applied during training to further enhance the model's generalization capability.





\subsection{Quantitative Results}
We compare the proposed RPCASSM with a range of recent state-of-the-art methods, including ACM\cite{dai2021asymmetric}, ALCNet\cite{dai2021attentional}, DNANet\cite{li2022dense}, AGPCNet\cite{zhang2021agpcnet}, UIUNet\cite{wu2022uiu}, MSHNet\cite{liu2024infrared}, RPCANet\cite{wu2024rpcanet}, SCTransNet\cite{yuan2024sctransnet}, LSDSSMs\cite{lu2025lsdssms}, DRPCA-Net\cite{xiong2025drpca}. 
These comparisons are performed on two public datasets: NUDT-SIRST and IRSTD-1K, where all models are trained according to a standard protocol.

\noindent \textbf{The quantitative experimental results} of different methods on NUDT-SIRST and IRSTD-1K datasets shown in \cref{tab:compare}. 
From the perspective of overall performance, our proposed RPCASSM achieves optimal or suboptimal performance on the vast majority of key indicators, showing excellent comprehensive performance.
Specifically, on the NUDT-SIRST dataset, our method ranked first in all three metrics of mIoU ($95.98\%$), Pd ($98.62\%$), Fa ($2.22\times 10^{-6}$) and F-measure ($97.95\%$). 
On the IRSTD-1K dataset, RPCASSM also performed well, ranking first in mIoU (68.44\%), Pd (92.44\%), and F-measure (81.26\%). Fa ($22.31\times 10^{-6}$) was also the lowest among all the methods. 
The excellent detection ability and low false alarm characteristics of the model are further verified.
In terms of model complexity, RPCASSM only contains $0.45M$ parameters, which is slightly higher than LSDSSMs ($0.37M$) and ALCNet ($0.42M$) with the least number of parameters, but still significantly lower than most comparison methods, achieving a good balance between accuracy and efficiency. 
In summary, RPCASSM not only achieves the leading detection and segmentation accuracy on the two big datasets, but also maintains a low number of model parameters, which proves the effectiveness and practicality of the proposed method.


\noindent \textbf{The ablation experimental results} shown in \cref{tab:ablation} clearly demonstrate the contribution of each module to the model performance. 
We take the baseline model as a starting point and gradually introduce the proposed module. 
First, adding DifBlock alone has led to consistent but weak improvements across all metrics, indicating its underlying effectiveness. 
Subsequently, the introduction of TSSM constitutes a key step, which significantly reduces Fa to 5.05 while continuing to improve mIoU and F-measure, which verfies the excellent ability of TSSM in focusing on the true target and suppressing false alarms. 
Finally, when BSSM is fully introduced, our model achieves the best comprehensive performance, with its mIoU, Pd and F-measure reaching 95.98\%, 98.62\%and 97.95\%, respectively, and Fa further decreasing to 2.22. 
The results fully prove the rationality and synergy of the design of our proposed dual-path state space architecture (TSSM and BSSM), which jointly ensure the complete modeling process from background suppression to target enhancement for infrared small targets, and finally achieve accurate and robust segmentation performance.

\noindent \textbf{The hyperparameter experimental results} are shown in Table \ref{tab:hyperparameters}, which show in detail the specific effects of different hyperparameter Settings on the performance of the model on the NUDT-SIRST dataset. 
We deeply analyze three key hyperparameters: the number of stages $k$, the weight $\alpha$ of the $\mathcal{L}_{IoU}(\cdot,\cdot)$, and the initialization threshold $T_{init}$. 
The experimental results show that when k=3, $\alpha$=0.01 and $T_{init}$ =0.7, the model can achieve the best performance, and the mIoU, Pd, Fa and F-measure reach 95.98\%, 98.62\%, 2.22 and 97.95\%, respectively. 
In addition, the experimental data also show that as the number of stages k increases, the performance of the model peaks at k=3, followed by a slight decrease at k=4 and k=5. 
The weight $\alpha$ is set to 0.1, which is more beneficial to improve the model performance. 
However, for the initialization threshold $T_{init}$, the model performs best when it is set to 0.7.


\noindent \textbf{The quantitative experimental results} of different methods on NUDT-SIRST and IRSTD-1K datasets are shown in \cref{tab:ROC}.  
From the perspective of overall performance, our proposed RPCASSM achieves optimal or suboptimal performance on the majority of evaluation metrics, demonstrating strong comprehensive capability.  
Specifically, on the NUDT-SIRST dataset, RPCASSM ranks first in both $\mathrm{AUC_{FPR=0.5 \times 10^{-4}}}$ (0.9340) and $\mathrm{AUC_{FPR=1.0 \times 10^{-4}}}$ (0.9529), significantly outperforming other methods.  
On the IRSTD-1K dataset, RPCASSM achieves the highest $\mathrm{AUC_{FPR=1.0 \times 10^{-4}}}$ (0.6916) and ranks second in $\mathrm{AUC_{FPR=0.5 \times 10^{-4}}}$ (0.5371), closely following MSHNet (0.5683).  
These results further verify the model’s excellent detection ability and low false alarm characteristics.  
In summary, RPCASSM achieves leading performance across both datasets while maintaining a compact model size, proving the effectiveness and practicality of the proposed method.

\subsection{Visual Comparisons}

The visualization results of \cref{fig:PCASSM.VISIMG} and \cref{fig:PCASSM.VIS3D} clearly reveal the performance differences of different infrared small target detection methods. In general, in the examples shown, only SCTransNet and the proposed method have no obvious missed detection or false alarm phenomenon, while other methods have different degrees of detection errors, including target missed detection, background false alarm and edge blurring. Especially in the complex scenes of the second, third, and fourth rows, which contain complex cloud background, low signal-to-noise ratio conditions, and low contrast between the target and the background, the proposed method shows better target edge modeling ability than SCTransNet, which can restore the target contour more completely and accurately, avoiding the loss and distortion of edge information. Thus, a clearer and more complete structure detection result is visually presented, which further verifies the advantage of the proposed method in detail preservation.

As shown in \cref{fig:ROC}, we compare the receiver operation characteristic(ROC) curves of different competitive algorithms. 
Analysis shows that on the NUDT-SIRST dataset, the proposed method maintains the highest detection rate in the overall false alarm rate range, showing comprehensive and stable performance advantages. 
On the IRSTD-1K dataset, although the performance is relatively flat in the low false alarm rate interval, with the increase of false alarm rate, the detection rate of the proposed method increases rapidly, and its curve finally surpasses that of most comparison methods, showing good scalability and asymptotic performance. 
Taken together, our proposed method shows more competitive detection performance on both datasets.



As shown in \cref{fig:TARGET}, the visualization analysis of the decomposed components $T_k$ from the RPCASSM model reveals that the target region exhibits distinct heterogeneous mutation signal characteristics along the single row axis.
The results demonstrate that the proposed model can progressively enhance the saliency of the target while suppressing the background interference.
Specifically, as the decomposition evolves from $T_1$ to $T_3$, the model effectively separates the low-rank background from the sparse target.
This process clearly illustrates the enhancement of heterogeneous signals, where the target peaks in the one-dimensional profile become increasingly prominent and well-defined, thereby validating the model's capability in highlighting subtle variations.

\section{Conclusion}
\label{sec:conclusion}

This paper proposes the RPCASSM network to address the limitation of existing mainstream visual state space models in accurately modeling the edge structures of infrared small targets. Unlike prior approaches that directly adopt generic visual state space architectures, our method is built upon the model paradigm of robust principal component analysis (RPCA) and leverages two intrinsic structural priors of infrared small targets: spatial saliency and target sparsity. Specifically, we design a Background State Space Module (BSSM) that introduces a spatial probe scanning mechanism to effectively model heterogeneous background information, and a Target State Space Module (TSSM) that employs a deformable prompt scanning mechanism guided by sparse cues to achieve fine-grained modeling of key target regions. Experimental results on two public datasets demonstrate that RPCASSM achieves superior performance in both detection accuracy and false alarm control, while maintaining a compact model size. These results validate the effectiveness of our design and highlight the importance of incorporating task-specific structural priors into state space models for infrared small target detection.

\bibliography{ref}
\bibliographystyle{IEEEtran}

\section{Biography Section}

\vspace{11pt}
\begin{IEEEbiography}[{\includegraphics[width=1in,height=1.25in,clip,keepaspectratio]{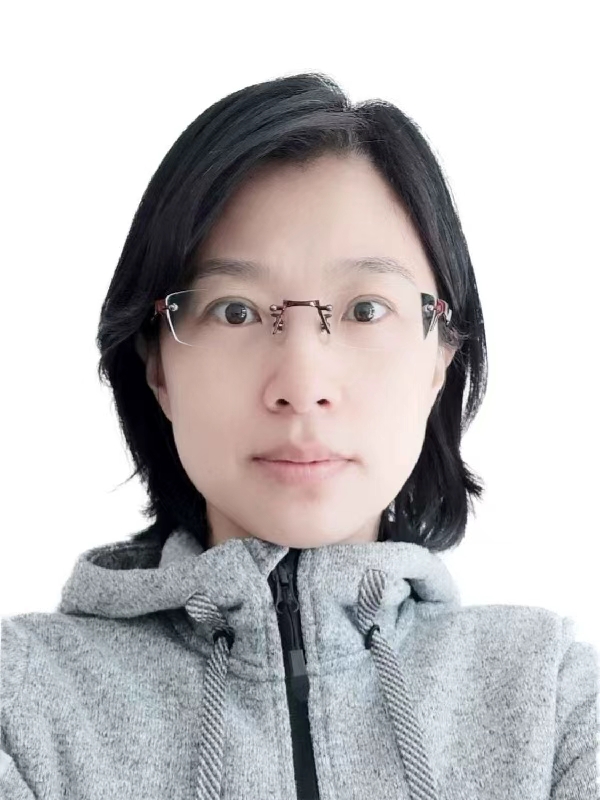}}]{Pingping Liu}
received M.S. and Ph.D. degrees from College of Computer Science and Technology,
Jilin University, China, in 2004 and 2009, respectively. She is currently a professor at College of
Computer Science and Technology, Jilin University. Her research interests include metric learning, image enhancement and restoration, and semi-supervised learning.
\end{IEEEbiography}

\vspace{11pt}
\begin{IEEEbiography}[{\includegraphics[width=1in,height=1.25in,clip,keepaspectratio]{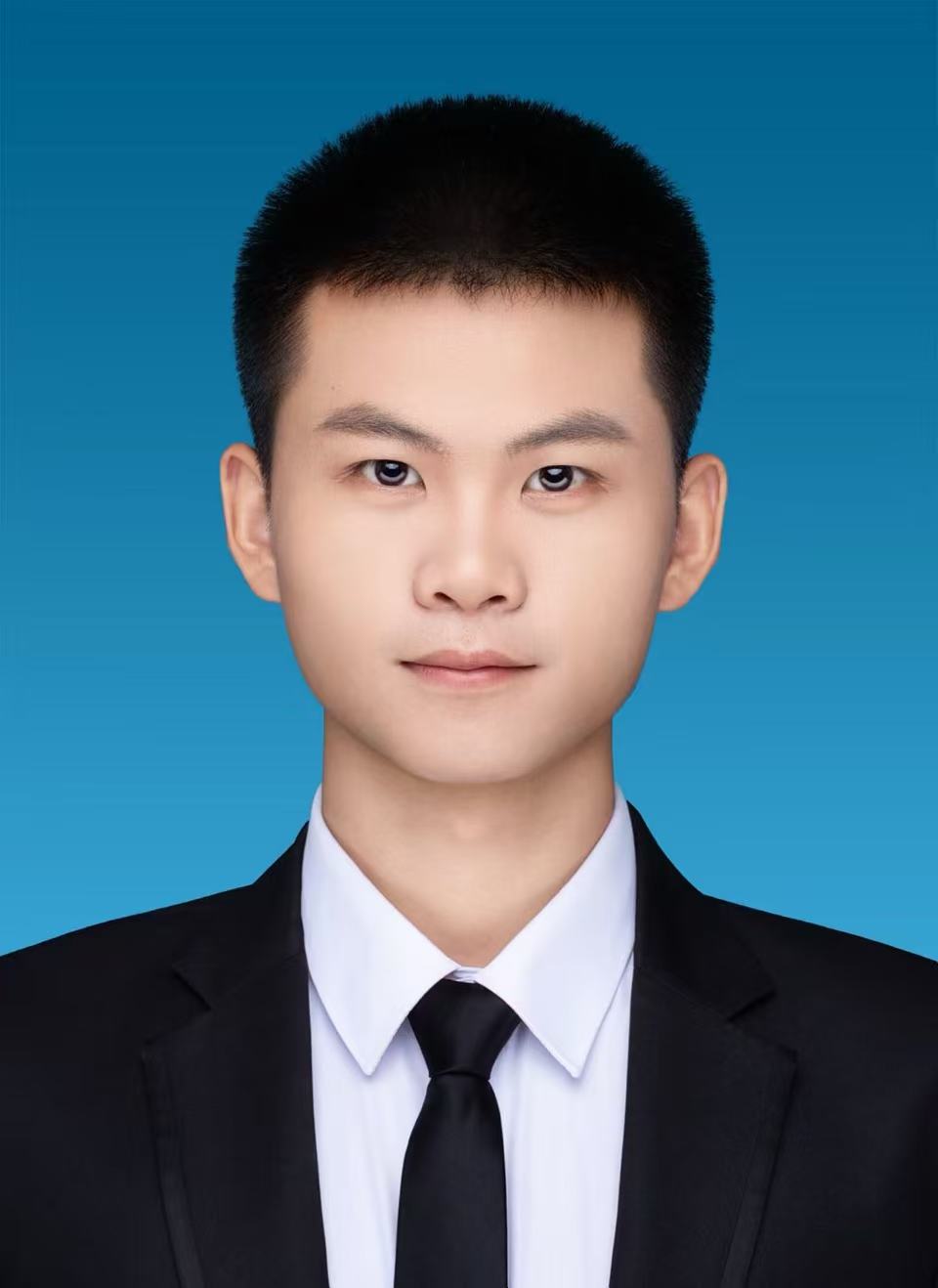}}]{Aohua Li}
was born in 2000. He received the B.S. degree from Hunan University of Technology in 2024. He is currently pursuing the M.S. degree with the College of Software, Jilin University, Changchun, China, where he focuses on research in machine vision and intelligent control. His research interests include domain adaptation, image segmentation and reinforce learning, with an emphasis on practical applications in intelligent systems.
\end{IEEEbiography}

\vspace{11pt}
\begin{IEEEbiography}[{\includegraphics[width=1in,height=1.25in,clip,keepaspectratio]{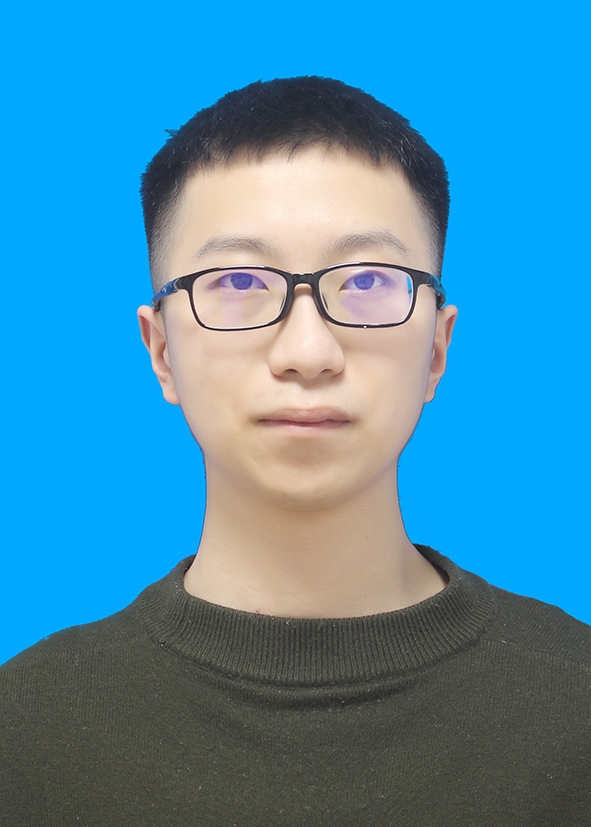}}]{Yubing Lu}
was born in 1997. He received the M.S. degree from Lanzhou University of Technology in 2023. He is currently pursuing his Ph.D. degree in College of Computer Science and Technology, Jilin University, China. His research interests include infrared small target detection, tracking, and image segmentation.
\end{IEEEbiography}

\vspace{11pt}
\begin{IEEEbiography}[{\includegraphics[width=1in,height=1.25in,clip,keepaspectratio]{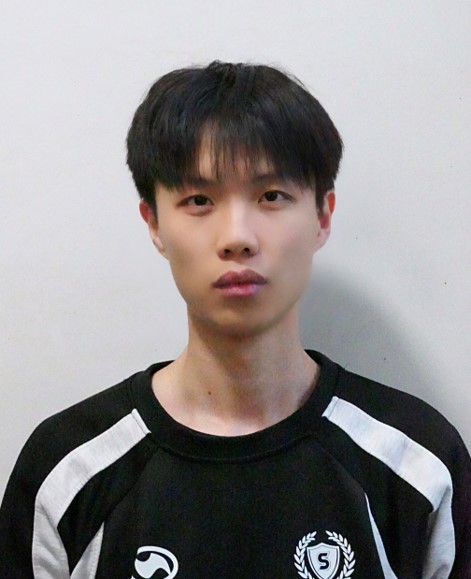}}]{Jin Kuang}
was born in 2001. He received the B.S. degree from Xiangnan University in 2022. He is currently pursuing the M.S. degree with Yangtze University, Wuhan, China, and serves as a Research Assistant with the Hunan Engineering Research Center of Advanced Embedded Computing and Intelligent Medical Systems. His research interests include domain adaptation, image segmentation, and low-light image enhancement.
\end{IEEEbiography}

\vspace{11pt}
\begin{IEEEbiography}[{\includegraphics[width=1in,height=1.25in,clip,keepaspectratio]{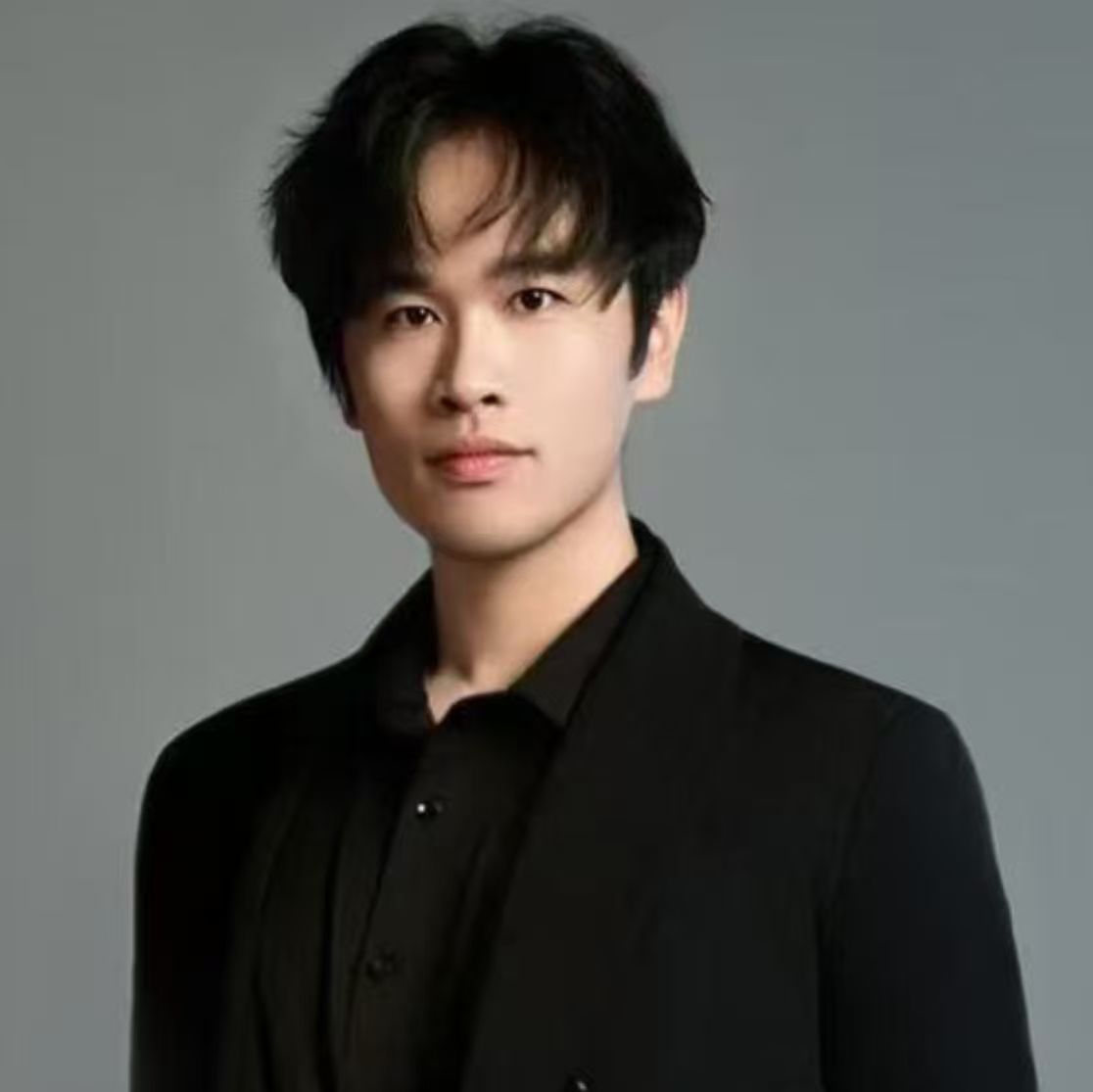}}]{Tongshun Zhang}
was born in 1996 in Shandong, China. He received the M.S. degree from Tiangong University in 2021. He is currently pursuing his Ph.D. degree in College of Computer Science and Technology, Jilin University, China. His research interests include low-light image enhancement, image inpainting, and image restoration.
\end{IEEEbiography}

\vspace{11pt}
\begin{IEEEbiography}[{\includegraphics[width=1in,height=1.25in,clip,keepaspectratio]{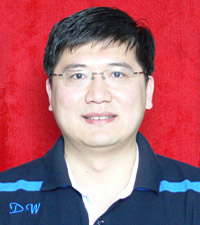}}]{Qiuzhan Zhou}
received M.S. and Ph.D. degrees from College of Communication Engineering, Jilin
University, China, in 2001 and 2004, respectively. He is currently a professor at College of Communication Engineering, Jilin University. His research interests include intelligent signal processing, recognition and data analysis.
\end{IEEEbiography}

\end{document}